\documentclass[sigconf,10pt]{acmart}

\usepackage{algorithm}
\usepackage{algpseudocode}
\usepackage{multirow}
\usepackage{enumitem}
\usepackage{float}
\usepackage{graphicx}
\usepackage{mathrsfs}
\usepackage{amsmath}
\usepackage{bm}
\usepackage{subcaption}
\usepackage{tabularx}
\usepackage{multirow}
\usepackage{booktabs}
\usepackage{xspace}

\usepackage{array}
\newcolumntype{P}[1]{>{\centering\arraybackslash}p{#1}}
\newcolumntype{M}[1]{>{\centering\arraybackslash}m{#1}}

\newcommand{\systemname}{\texttt{sd.npu\xspace}}

\newcommand{\modified}[1]{{#1}}

\renewcommand\footnotetextcopyrightpermission[1]{}
\begin{document}

\title[Paper]{Accelerating Mobile Language Model via Speculative Decoding and NPU-Coordinated Execution}

\settopmatter{authorsperrow=0}
\author{Zhiyang Chen}
\orcid{0009-0006-8607-8539}
\affiliation{%
  \institution{Peking University}
  \country{China}
}

\author{Daliang Xu}
\orcid{}
\affiliation{%
  \institution{Beijing University of Posts and Telecommunications}
  \country{China}
}
\authornote{Corresponding authors.}

\author{Haiyang Shen}
\orcid{0009-0000-4599-3198}
\affiliation{%
  \institution{Peking University}
  \country{China}
}

\author{Chiheng Lou}
\affiliation{%
  \institution{Peking University}
  \country{China}
}

\author{Mengwei Xu}
\orcid{}
\affiliation{%
  \institution{Beijing University of Posts and Telecommunications}
  \country{China}
}

\author{Shangguang Wang}
\orcid{}
\affiliation{%
  \institution{Beijing University of Posts and Telecommunications}
  \country{China}
}

\author{Xin Jin}
\affiliation{%
  \institution{Peking University}
  \country{China}
}

\author{Yun Ma}
\orcid{0000-0001-7866-4075}
\affiliation{%
  \institution{Peking University}
  \country{China}
}

\authornotemark[1]

\renewcommand{\shortauthors}{Chen et al.}

\begin{abstract}
Performing Retrieval-Augmented Generation (RAG) directly on mobile devices is promising for data privacy and responsiveness but is hindered by the architectural constraints of mobile NPUs. Specifically, current hardware struggles with the variable workloads intrinsic to RAG: the transition between processing extensive contexts and generating tokens incurs significant overhead due to static graph constraints, while the memory-bound generation phase leaves computational resources underutilized.

In this work, we propose a holistic acceleration framework \systemname, designed to maximize NPU efficiency for on-device RAG ecosystem. To address the latency caused by NPU graph switching during phase transitions, we introduce a pipelined execution strategy. This approach masks the overhead of model reconfiguration by parallelizing the loading of decoding graphs with the computation of partitioned context chunks (chunked prefill), thereby ensuring continuous execution flow. Furthermore, to mitigate low hardware utilization during the decoding phase, we develop an NPU-centric speculative decoding mechanism. By calibrating generation distributions and extending draft sequences, our method effectively converts idle NPU cycles into valid token throughput. Experiments on commercial smartphones show that our framework significantly outperforms existing baselines, delivering 1.06$\times$--3.81$\times$ speedups and 1.07$\times$--4.71$\times$ energy savings across various RAG tasks.
\end{abstract}




\maketitle

\section{Introduction} \label{sec:introduction}

Retrieval-augmented generation (RAG) has become the de facto standard for grounding Large Language Model (LLM) outputs in user-specific data, enabling a new class of edge applications ranging from personalized document assistants to automated UI agents~\cite{usecase:ui_automation, work:mobile_rag1, work:mobile_rag2, work:mobile_rag3}. As these applications increasingly handle sensitive personal data, RAG workloads are shifting towards \emph{on-device} execution to ensure privacy and reduce cloud reliance~\cite{trend:slm1, trend:slm2, privacy1, privacy2, privacy3}. In this setting, the retrieval phase is typically fast (e.g., $<10$\,ms for local corpora~\cite{work:edge_rag, work:webanns}), leaving the computational heavy lifting to the LLM generation phase.

To sustain always-on RAG capabilities without draining the battery or stalling the user interface, the Neural Processing Unit (NPU) has emerged as the critical hardware accelerator. Unlike CPUs and GPUs that are burdened with OS services, application logic, and graphics rendering, NPUs are domain-specific architectures dedicated to tensor operations. They offer two distinct advantages: (i) superior energy efficiency, delivering significantly lower energy-per-token for Transformer workloads compared to general-purpose processors~\cite{work:mllm, work:heterollm, work:tman}; and (ii) performance isolation, allowing background RAG inference to run concurrently with foreground tasks without contending for CPU/GPU resources~\cite{work:npu_sparse_attn}. Driven by these benefits, recent systems have moved towards \emph{NPU-only} inference~\cite{work:tman, work:ttc, work:npu_sparse_attn}, executing the entire pipeline on the NPU to eliminate the high synchronization and data movement costs inherent in heterogeneous CPU-NPU co-scheduling~\cite{work:mllm, work:heterollm}.


Despite these advantages, NPU-only on-device RAG inference still suffers from long decoding latency, driven by two critical challenges:


\noindent $\bullet$ \textbf{Static graph architecture vs. dynamic RAG phases.}
Mobile NPUs rely on static compute graphs with fixed tensor shapes, buffer layouts, and execution plans. Yet RAG inference alternates between two computational extremes: \emph{prefill}, which processes long, context-heavy prompts (suitable graph for 128/256 tokens input), and \emph{decoding}, which autoregressively generates tokens one-by-one (suitable graph for 1 token input). The gap between these two regimes is too large that no single static graph can efficiently serve both. To fit within tight on-device memory budgets, existing systems keep only one graph resident at a time,  necessitating an expensive teardown-and-load process when switching phases: unloading the large prefill graph and loading the decoding graph.
We find that naive graph switching accounts for up to 29\% of the decoding latency in on-device RAG settings.

\noindent $\bullet$  \textbf{Matrix-oriented hardware vs. memory-bound decoding.}
While prefill saturates the NPU's matrix units, decoding is memory-bound, leaving compute resources severely underutilized.
Speculative Decoding (SD)~\cite{work:sd_is_lossless} is a standard remedy that enables parallel decoding through a draft-then-verify schema, but existing model-based SD methods (e.g., EAGLE~\cite{work:eagle}) require auxiliary models, leading to high mobile memory consumption and slow speculative inference on NPUs.
For RAG specifically, \textit{retrieval-based SD}~\cite{work:sam} is theoretically ideal: it drafts candidate tokens directly from context by first-token retrieval matching without requiring the extra cost of auxiliary draft models.
\emph{However, we observe that off-the-shelf retrieval-based SD fails to saturate mobile NPUs due to two inefficiencies.} 
To understand why, we introduce the concept of effective NPU utilization and decompose the NPU cycles spent during verification into three categories: (i) \emph{Valid}: cycles spent verifying tokens that are eventually accepted; (ii) \emph{Wasted}: cycles spent verifying tokens that are rejected; and (iii) \emph{Underutilized}: idle capacity in the matrix engines when the workload is too small (low batch size/sequence length). Our analysis (§~\ref{sec:motivation}) shows that existing approaches are dominated by the following inefficiencies:
\begin{enumerate}[noitemsep, leftmargin=*, topsep=0pt, align=left, labelsep=-3pt, label=(\roman*)]
    \item \emph{Lexical Divergence causes Wasted Compute:}
    retrieved context often matches the semantic intent but differs lexically from the LLM’s training distribution. This causes the verifier to reject most draft tokens. In our measurements, \textit{22\%} of NPU cycles are wasted verifying drafts that are ultimately rejected.
    \item \emph{Short Drafts cause Underutilization:} 
    NPU matrix units (e.g., Qualcomm HMX~\cite{tool:qnn_sdk}, Huawei Cube~\cite{tool:ascend}) rely on large operational intensity (large batch or sequence length) to hide memory latency and satisfy the minimal tile size (e.g., 32 tokens for HMX). However, retrieval-based drafts are often fragmented and short. We find that over 75\% of drafts are shorter than 8 tokens, whereas mobile NPUs typically require sequences of 32+ tokens to approach peak utilization. This mismatch leaves more than 70\% of the NPU's compute capacity underutilized during verification steps.
\end{enumerate}

\noindent \textbf{Design} We present \systemname, the first system designed to enable efficient, end-to-end NPU offloading for on-device RAG. It bridges the gap between the NPU's requirement for static, massive parallelism and the dynamic-shape, fragmented nature of RAG workloads by incorporating several techniques:

\noindent $\bullet$ \textbf{Progressive graph scheduling (§~\ref{sec:tech1})} addresses the \textit{monolithic loading stall} caused by deferring graph switching to the final prefill stage. Instead of a sudden, blocking switch, we amortize the heavy I/O cost across the entire prefill phase by leveraging two key insights.
(i) \textit{Decomposition}: the monolithic LLM is partitioned into fine-grained blocks (subsets of Transformer layers), making the I/O payload of a single block small enough to be masked by several chunked prefill computation.
(ii) \textit{Equivalence}: we observe that repeated, iterative execution of a decoding-friendly block graph is mathematically equivalent to a single execution of a prefill-optimized graph.
Based on these, \systemname~orchestrates a multi-iteration progressive transition. Specifically, by partitioning the LLM, we progressively orchestrates the block-by-block switching of the prefill graphs to their decoding-friendly counterparts; each block switching is meticulously designed to be perfectly overlapped with one chunked prefill operation. This ensures a seamless and zero-cost transition to the decoding phase. After the decoding graph for each block has been loaded, it is executed iteratively to process the corresponding chunk of the prefill input, thereby ensuring functional equivalence and maintaining the correct overall prefill output.

\noindent $\bullet$ \textbf{NPU-tailored speculative decoding (§~\ref{sec:tech2}, \ref{sec:tech3})}. Building upon retrieval-based speculative decoding, \systemname~transforms the "wasted" and "underutilized" NPU cycles into "valid" generation throughput, incorporating two distinct techniques:

\begin{enumerate}[noitemsep, leftmargin=*, topsep=0pt, align=left, labelsep=-3pt, label=(\roman*)]
\item  \textit{In-context distribution calibration (§~\ref{sec:tech2}).} To minimize \emph{wasted} compute caused by lexical divergence, we must align the retrieved drafts with the model's output distribution. Cloud-based solutions might perform offline fine-tuning on the knowledge base, but this is infeasible on edge devices due to battery, storage, and privacy constraints. Instead, \systemname~adopts a zero-cost, online calibration. We observe that during the prefill, the model inherently computes the logits for the provided context. \systemname~reuses these existing logits to construct a model-calibrated token tree. By drafting from this calibrated tree rather than the raw text, we reduce draft rejection without any additional computation or model training.
\item  \textit{NPU-optimized draft extension (§~\ref{sec:tech3}).} Minimizing \emph{underutilized} compute caused by short drafts requires longer draft sequences that can saturate the NPU matrix units. Existing SD methods focus solely on the \emph{input} side (drafting) and discard rejected tokens. However, we observe that rejection is often temporary: around 38.5\% of tokens rejected in one step become valid in subsequent steps as the context evolves. Based on this insight, \systemname~implements a verification-aware draft extension. It recycles high-confidence rejected tokens from previous steps to extend the current draft. This not only recovers useful information but also artificially lengthens the verification batch, pushing the NPU into a higher-utilization regime.
\end{enumerate}





\noindent \textbf{Evaluation} We implement \systemname~on top of \texttt{mllm}~\cite{work:mllm}, an open-source mobile LLM inference system with NPU support, and evaluate it across three smartphones (Redmi K60 Pro, Redmi K70 Pro, OnePlus 13), four datasets (summarization, QA, UI automation, auto-reply), and three models (Qwen2.5-1.5B, Qwen2.5-0.5B, LLaMA3.2-3B). 
Results show that \systemname~effectively eliminates graph switching costs and transforms 70.5\% wasted and underutilized NPU computation into valid ones. \systemname~achieves 1.06--3.81$\times$ end-to-end speedup and 1.11--4.18$\times$ energy reduction compared to the vanilla system, while outperforming existing speculative decoding integrations by up to 2.53$\times$ in latency and 4.71$\times$ in energy. 
Our contributions are summarized as follows:

\noindent $\bullet$ Identify and characterize the key inefficiencies of end-to-end NPU offloading for RAG workloads, revealing optimization opportunities via speculative decoding.  

\noindent $\bullet$ Design and implement \systemname, a system that co-optimizes speculative decoding and NPU execution through two novel techniques, bridging the gap between compute-bound LLM decoding and NPU offloading.  

\noindent $\bullet$ Conduct comprehensive evaluations across multiple devices, models, and tasks, demonstrating consistent latency and energy efficiency improvements in diverse edge environments.  

\section{Background}
\label{sec:background}

\subsection{RAG for Mobile Applications}
\label{sec:background:sub1}

Retrieval‑augmented generation (RAG) has become an essential technique for on‑device intelligent applications~\cite{work:rag, work:mobile_rag1, work:mobile_rag2}, such as mobile UI agents and personal assistants. By retrieving relevant external documents (e.g., UI information for automation and user archives for personalization) and injecting it into the prompt, RAG improves the accuracy and reliability of LLM‑driven tasks. Compared to cloud‑based inference, on‑device RAG offers strong privacy guarantees and consistent responsiveness, making it well suited for scenarios involving sensitive data or real‑time interactions~\cite{work:adaptive_rag}. 
Recent efforts on retrieval acceleration have largely addressed its overhead, achieving latency of several microseconds for querying on GB-size corpus~\cite{work:webanns, work:edge_rag}. Thus, our focus is the compute burden introduced by LLM inference, as the augmented context significantly lengthens prompts, amplifying the cost of both prefill and decoding~\cite{work:rag_amplify_cost1, work:rag_amplify_cost2}. Fortunately, modern mobile SoCs are increasingly integrating neural processor units (NPUs) with high compute capacity, offering opportunities for hardware acceleration.

\subsection{Mobile NPUs}
\label{sec:background:sub2}

\begin{figure}[!t]
  \centering
\includegraphics[width=1.0\linewidth]{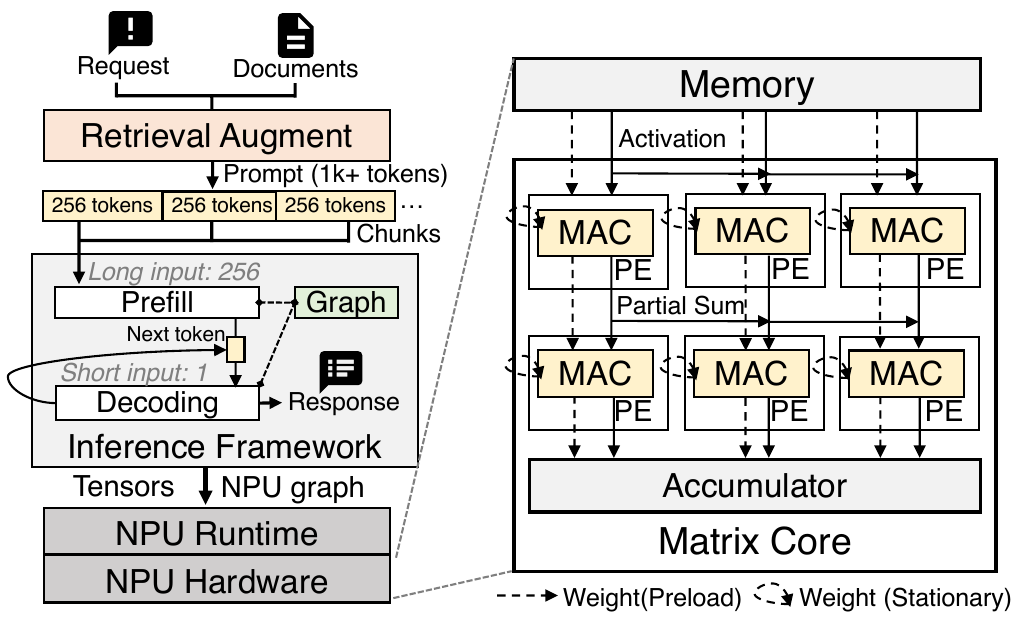}
  \caption{RAG on NPU and NPU's architecture.
  }
  \label{fig:npu_arch}
  \vspace{-1em}
\end{figure}

Modern mobile NPUs are specialized accelerators designed to deliver high throughput and energy‑efficient execution for deep neural networks. Their compute cores are typically built around matrix computation units, which usually are implemented as systolic arrays with weight‑stationary designs~\cite{device:8gen3, device:8gen4}. As illustrated in Figure~\ref{fig:npu_arch}, systolic arrays preload weight tiles into their processing elements and keep them stationary during computation. This weight‑stall paradigm amortizes the cost of weight movement and maximizes reuse across input activations, achieving high efficiency for large matrix multiplications. 
Meanwhile, the fixed systolic array constraints the minimal input dimensions to effectively saturate the processing units, as each tile in the array (e.g., 32×32 FP16 or INT8) corresponds to the fundamental execution unit of the NPU. To address this, the NPU runtime compiles operators into sequences of tile‑level matmul instructions, pads and partitions input tensors into tiles such that activations could be streamed through the systolic array with the optimal efficiency.
Therefore, most mobile NPU SDKs require neural networks to be converted into static compute graphs with fixed tensor shapes, memory bindings, and operator layouts. Graph compilation can take several seconds and is generally performed offline~\cite{work:mllm}. At runtime, the NPU executes these static graphs with minimal control‑flow support, which simplifies scheduling overhead by restricting dynamic shape changes and conditional execution.

\subsection{Accelerating RAG on Mobile NPUs}
\label{sec:background:sub3}



\begin{table}[t]
\centering
\footnotesize
\begin{tabular}{l l l r}
\hline
Vendor & SoC & NPU & Perf. (Tops) \\
\hline
Qualcomm & 8 Gen 3   & Hexagon NPU   & 73 \\
Apple    & A18       & Neural Engine & 35 \\
MediaTek & K9300     & APU 790       & 60 \\
Huawei   & Kirin-9000& Ascend NPU    & 16 \\
\hline
\end{tabular}
\begin{flushleft}
\footnotesize{Perf. = INT8 Performance in Tops.}
\end{flushleft}
\caption{Specifications of representative mobile NPUs.  }
\label{tab:npu_specs}
\vspace{-1em}
\end{table}


    
    
    
    



As mobile SoCs increasingly integrate NPUs (Table~\ref{tab:npu_specs}), leveraging these accelerators for RAG has become a critical optimization direction. When deploying RAG systems on edge devices, existing inference engines typically adopt one of two execution strategies:

\noindent $\bullet$ \textbf{NPU-CPU/GPU co-scheduling} (e.g., mllm~\cite{work:mllm}, HeteroLLM~\cite{work:heterollm}, PowerInfer‑V2~\cite{work:powerinfer2}): offloading only compute-intensive operations in prefill to NPU, while executing precision-intensive kernels and decoding on CPUs or GPUs.

\noindent $\bullet$ \textbf{NPU-only execution} (e.g., T-MAN~\cite{work:tman}, test-time-compute NPU~\cite{work:ttc} and ExecuTorch~\cite{work:executorch}): offloading all computation to NPU through carefully designed quantization strategies.

\begin{figure}[!t]
  \centering
\includegraphics[width=0.9\linewidth]{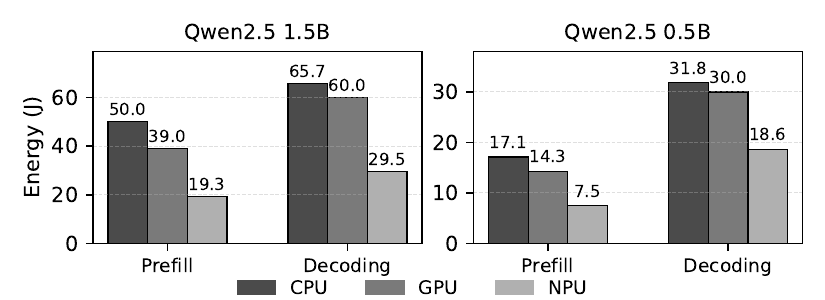}
  \caption{Comparison on consumed energy per request across different execution backends~\cite{work:mllm, work:heterollm}.
  }
  \label{fig:energy_compare}
  \vspace{-1em}
\end{figure}

Among these, the NPU-only approach is particularly promising for sustainable RAG acceleration.
As shown in Figure~\ref{fig:energy_compare}, NPU-only execution reduces energy consumption by eliminating frequent cross‑processor scheduling and by consolidating execution onto the most energy‑efficient compute unit~\cite{work:mllm, work:heterollm}. Furthermore, it also minimizes resource contention with other general-purpose mobile workloads processed by CPU and GPU, such as user interaction handling and rendering~\cite{work:npu_sparse_attn}. Therefore, this paper focuses on addressing the challenges within the NPU-only acceleration path.

However, realizing the full potential of NPU-only execution remains challenging. Prior work (e.g., test-time-compute NPU~\cite{work:ttc}, T-MAN~\cite{work:tman}) primarily relies on aggressive quantization strategies~\cite{work:decdec}, which necessitate invasive changes to NPU kernels and proprietary SDKs. Moving beyond kernel-centric approaches, we aim to optimize RAG performance through system-level innovations. Our optimizations address the scheduling and memory inefficiencies inherent in NPU runtimes and are designed to be orthogonal to specific quantization schemes, enabling additive performance gains atop these kernel-level baselines.


\section{Challenges of RAG on NPU}
\label{sec:motivation}

Figure~\ref{fig:npu_arch} depicts the NPU-based RAG workflow. Excluding the negligible retrieval overhead, execution comprises two distinct phases: \textit{compute-bound prefill} and \textit{memory-bound decoding}. During prefill, the variable-length context is partitioned into fixed-size chunks (e.g., 256 tokens) to accommodate the NPU's static shape requirement~\cite{work:mllm, work:heterollm}. However, this execution paradigm exposes two critical inefficiencies on resource-constrained mobile devices:
(i) graph switching vs. padding dilemma (§~\ref{sec:motivation_sub1}): due to strict memory constraints, mobile runtimes can typically reside only one active static graph. Given the drastic shape disparity between prefill (e.g., 256 tokens) and decoding (1 token), the system faces a prohibitive trade-off: either \textit{switch graphs}, which incurs high latency from model reloading, or \textit{reuse the prefill graph} via padding, which results in massive computation waste.
(ii) NPU underutilization in decoding (§~\ref{sec:motivation_sub2}): the decoding phase fails to saturate the NPU's massive matrix parallelism. 
A typical widely-used technique to alleviate this by increasing concurrency is speculative decoding (SD)~\cite{work:sd_is_lossless, work:eagle2, work:rest}. However, we find traditional SD still fails to saturate the NPU's systolic array architecture due to a mismatch with the hardware cost model.
Crucially, leaving NPU compute resources significantly idle compared to the dense prefill phase.


\subsection{Overhead of Static Graph Management}
\label{sec:motivation_sub1}

\begin{figure}[t]
  \centering  \includegraphics[width=1.0\linewidth]{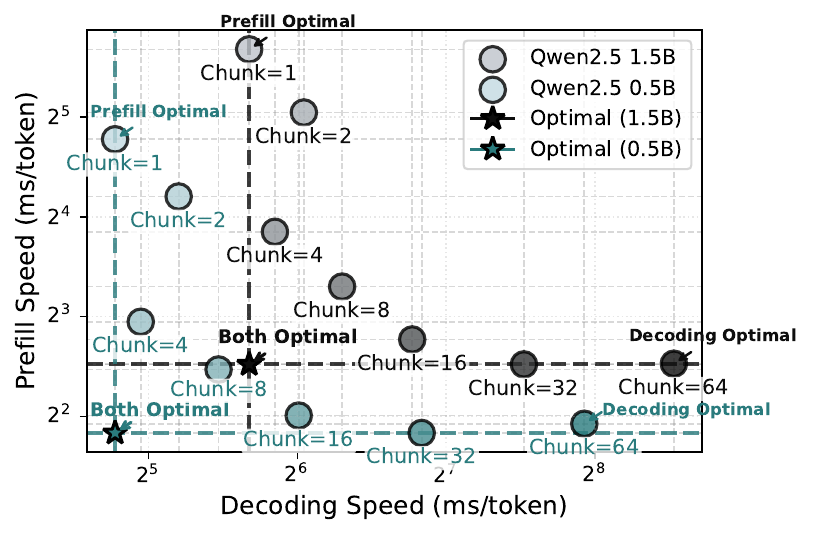}
  \caption{Trade-off between prefill and decoding with fixed-size compute graphs, highlighting the importance to deploy specialized graphs for each stage. 
  }
  \label{fig:npu_prefill_vs_decoding_comparison}
\end{figure}

The NPU's reliance on static compute graphs creates a fundamental tension between computational efficiency and runtime latency.
While large-tile graphs benefit the dense prefill phase, decoding requires latency-optimized, small-shape graphs.
As shown in Figure~\ref{fig:npu_prefill_vs_decoding_comparison}, this shape mismatch is critical: forcing the decoding phase to reuse a prefill-optimized graph (via padding) degrades throughput by up to 8$\times$ due to excessive redundant computation.

\begin{figure}[t]
  \centering  
  \includegraphics[width=1.0\linewidth]{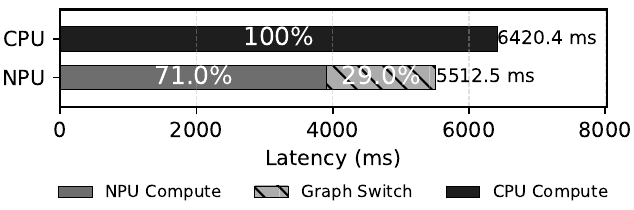}
  \caption{Latency breakdown of RAG decoding phase.}
  \label{fig:latency_breakdown_between_cpu_npu}
\end{figure}

Consequently, runtimes are compelled to perform graph switching, i.e., unloading the prefill graph and reloading a decoding-specialized graph. However, this operation is far from free.
Our breakdown of decoding latency (Figure~\ref{fig:latency_breakdown_between_cpu_npu}) reveals that graph switching alone accounts for 29.0\% of the total time for RAG tasks with 1K context. Naive switching blocks execution and saturates memory bandwidth, while keeping both graphs resident is infeasible due to the tight RAM budget of mobile devices.
In contrast, CPU decoding avoids this overhead but suffers from significantly lower raw compute capability.



\noindent \textbf{Challenge \#1}: \textit{the static graph dilemma.}
Efficient NPU offloading is deadlocked: reusing the prefill graph wastes computation (8$\times$ slowdown), while switching to a specialized graph incurs prohibitive latency (29\% overhead). This motivates the need for a runtime mechanism that minimizes graph switching costs without compromising efficiency.

\subsection{Ineffectiveness of Speculative Decoding on Mobile NPUs}
\label{sec:motivation_sub2}

\begin{figure}[t]
  \centering  \includegraphics[width=1.0\linewidth]{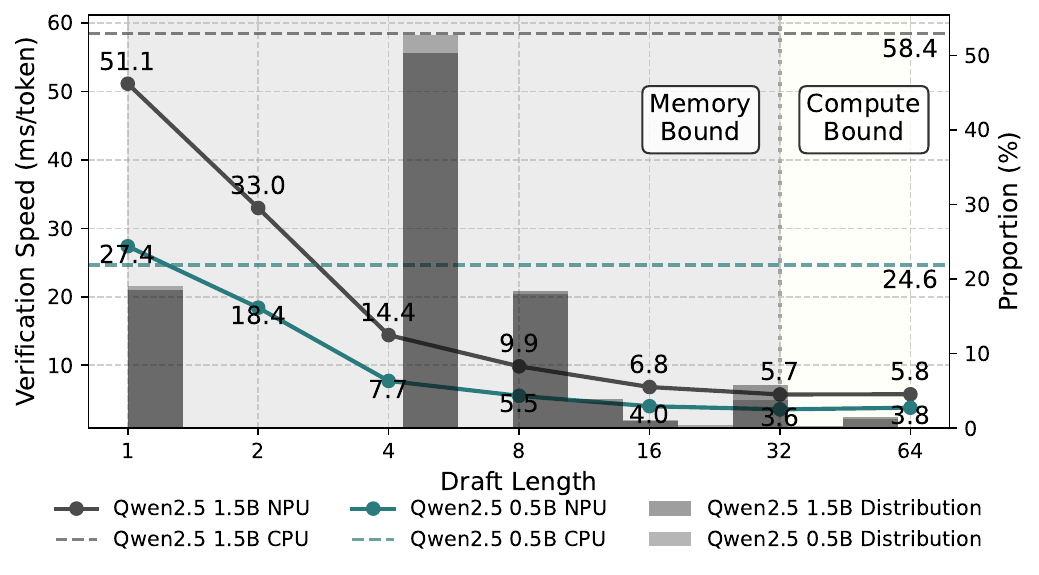}
  \caption{Comparison of verification speed with different draft lengths, and the length distribution of drafts generated by existing SD method~\cite{work:pld} (histogram). }
  \label{fig:cpu_npu_decoding_speed}
\end{figure}

\begin{figure}[t]
  \centering  
  \includegraphics[width=1.0\linewidth]{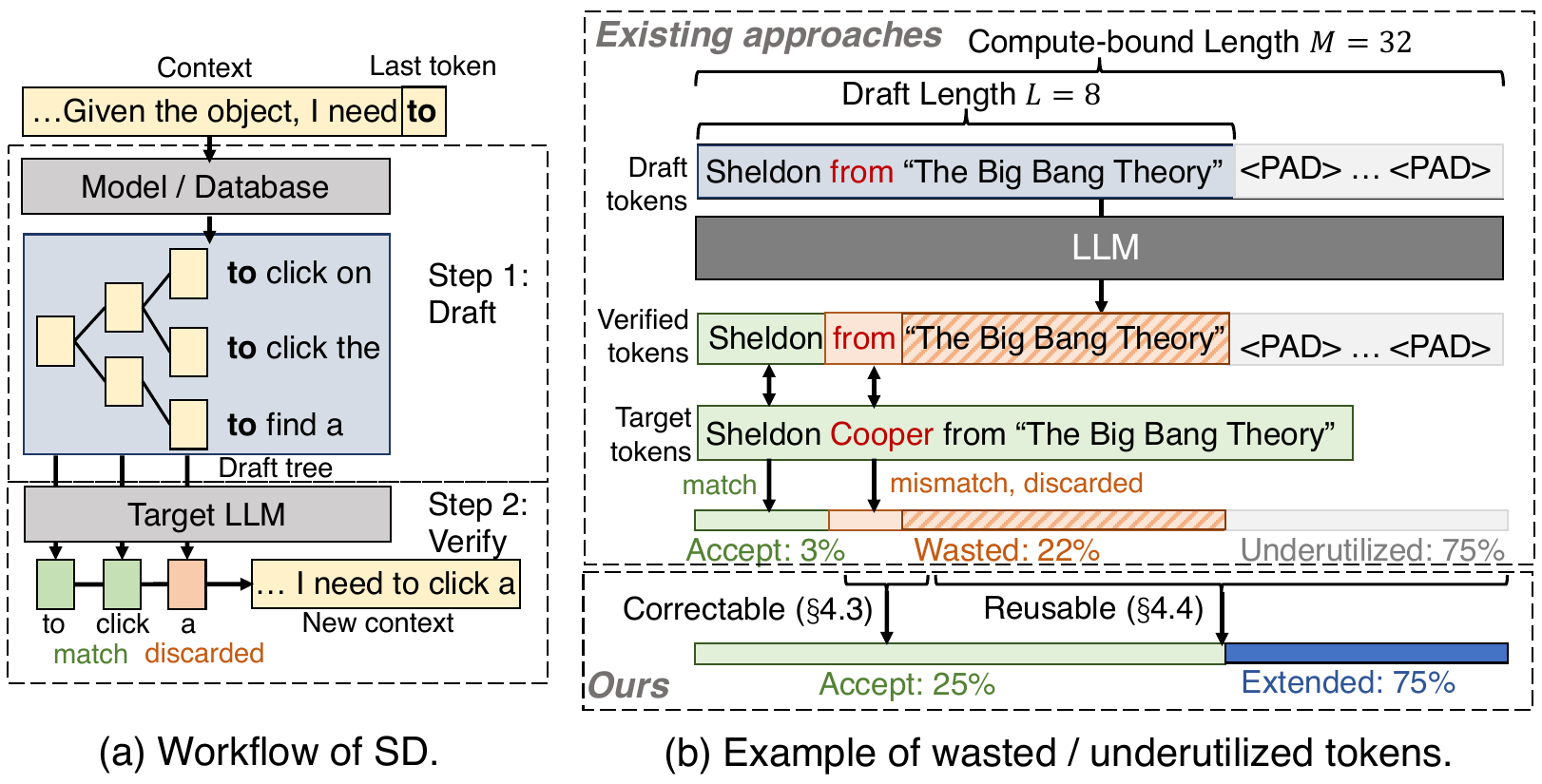}
  \caption{Workflow of speculative decoding and an example of existing inefficiencies: a large proportion of computation is wasted or underutilized.}
  \label{fig:sd_demo}
  \vspace{-4pt}
\end{figure}

Besides graph scheduling challenge, the NPU‑only decoding phase faces another bottleneck: memory bandwidth and NPU underutilization.
Generating a single token on the NPU is memory-bound due to low arithmetic intensity.
In this regime, the NPU's massive matrix array is starved, degrading throughput to CPU levels while consuming the fixed overhead of NPU kernel invocation.
To amortize these overheads and saturate the NPU, the workload must be transformed from sequential vector operations into parallel matrix operations.


Speculative decoding~\cite{work:sd_is_lossless} is a widely adopted method to effect this transformation. SD uses a lightweight model (i.e., model-based methods~\cite{work:medusa, work:eagle}) or database (i.e., retrieval-based methods~\cite{work:pld, work:rest}) to generate multiple draft tokens and verifies them in parallel. In verification, only correct tokens are accepted and any tokens after the first mismatch is discarded (Figure~\ref{fig:sd_demo}a). This draft-then-verify schema allows SD to convert memory-bound decoding into compute-bound verification while maintaining lossless generation quality~\cite{work:sd_is_lossless}. 
Among different approaches, model-based SD is infeasible for mobile devices as it requires a separate draft model (often $>1$GB), which competes for scarce RAM and compute capacity with the main LLM and the OS. Instead, for RAG workloads, the retrieved documents intrinsically contain the information needed for generation. Using the context as a draft source~\cite{work:sam} incurs negligible overhead, making it the ideal candidate for mobile RAG. Consequently, this paper adopts retrieval-based speculative decoding as the foundational mechanism for mobile acceleration.

\noindent \textbf{Breakdown of SD inefficiency.} Despite its theoretical suitability, we find that off-the-shelf algorithms fail to accelerate NPU inference. To understand this inefficiency, we decompose NPU computation during verification into three parts: (i) \textit{valid}, resources spent on accepted drafts; (ii) \textit{wasted}, resources spent on rejected drafts; and (iii) \textit{underutilized}, the NPU's idle capacity when the workload is too small. As show in Figure~\ref{fig:sd_demo}b, our measurements reveal that wasted (22\%) and underutilized portions (75\%) dominate the use of compute resources, leaving the effective throughput surprisingly low.

\begin{figure}[t]
  \centering  
  \includegraphics[width=0.9\linewidth]{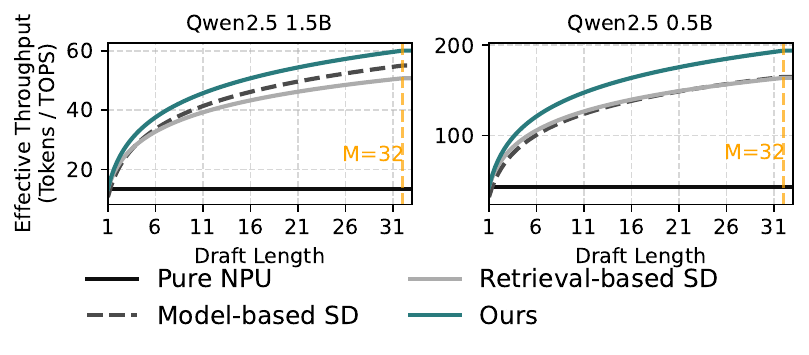}
  \caption{Effective throughput comparison. For NPU, throughput is dictated by accept length.}
  \label{fig:sd_compare}
  \vspace{-4pt}
\end{figure}

\noindent \textbf{Root cause: constant-cost trap.}
This inefficiency stems from the unique performance characteristics of mobile NPUs. Unlike GPUs where cost scales linearly, NPU verification latency remains nearly constant up to a hardware saturation point $M$ (Figure~\ref{fig:sd_compare}).
In this regime, decoding throughput is dictated solely by the total accepted length per step. To achieve speedup, the system must maximize the number of accepted tokens within this constant-time window.
However, existing retrieval-based methods fail to exploit this characteristic. For instance, as shown in §~\ref{sec:ablation}, PLD~\cite{work:pld} achieves an average accepted length of only 2.53 tokens per step, resulting in less speedup. This failure manifests in two specific ways:


\noindent \textbf{Challenge\#2.1:} \textit{wasted computation due to low draft acceptance}. To preserve lossless generation, SD rejects drafts that lexically mismatch the base model's predictions, even if they are semantically aligned. In RAG workloads, this lexical divergence is common, causing frequent rejections. Since the NPU incurs a near-constant high cost for each verification step regardless of the outcome, every rejection squanders a full pass of NPU computation. This directly contributes to a high percentage of wasted computation.

\noindent \textbf{Challenge\#2.2:} \textit{underutilized computation due to short workloads}. The NPU's constant-cost model means that peak efficiency is only achieved with long drafts that saturate its compute units. However, existing SD methods for RAG often produce very short drafts. Empirically, over 90\% of drafts contain fewer than 32 tokens (Figure~\ref{fig:cpu_npu_decoding_speed}). Verifying such short sequences leaves most of the NPU's matrix units idle, paying the high cost of a full verification step for only a few tokens. This severely throttles effective throughput, sometimes making NPU even slower than CPU.

Together, these observations motivate the need for an new SD algorithm explicitly co-designed with RAG workloads and NPU execution characteristics. 

\section{Methodology}

\subsection{Overview}
\label{sec:overview}

\begin{figure}[!t]
  \centering
\includegraphics[width=1.0\linewidth]{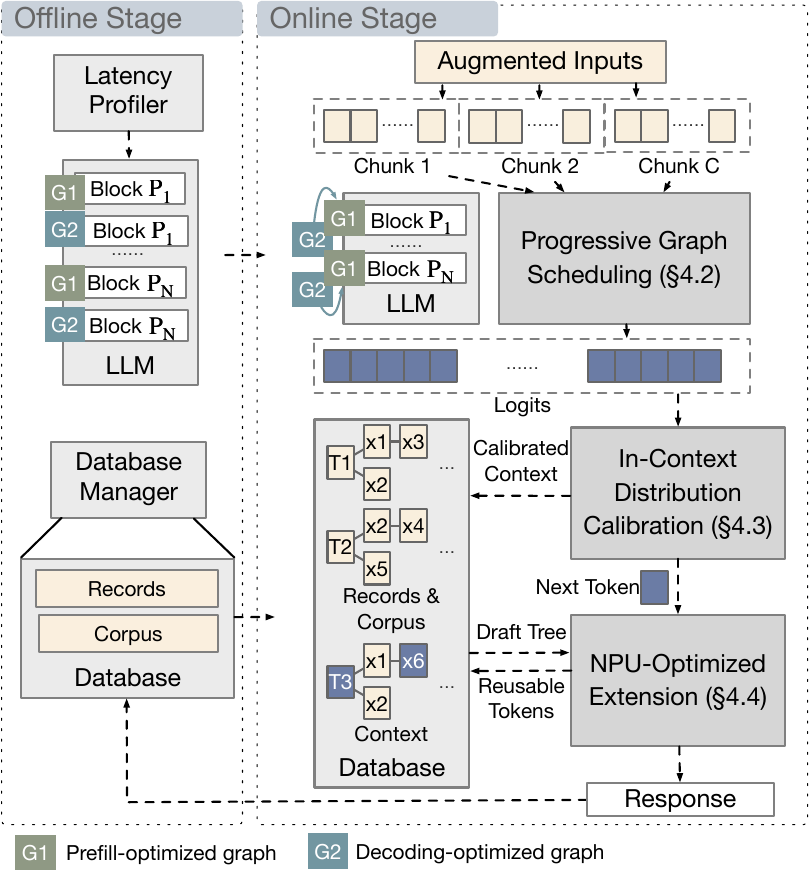}
  \caption{Overview of \systemname.}
  \label{fig:system_overview}
  \vspace{-4pt}
\end{figure}

\noindent \textbf{Design goal.}
To enable efficient RAG inference with end-to-end NPU offloading, \systemname~is designed as a runtime system that couples speculative decoding with NPU execution. Its key idea is a system–algorithm co-design: at the system level, \systemname~reduces graph-switching overhead, and at the algorithmic level, it introduces an NPU-optimized speculative decoding scheme.

\noindent \textbf{Architecture and workflow.}
Figure~\ref{fig:system_overview} illustrates the overall architecture.
It intercepts the standard prefill-decoding pipeline and orchestrates speculative decoding workloads tailored for NPU execution with the following workflow:

\noindent $\bullet$ \textit{Offline initialization.}
\systemname~partitions the LLM into several blocks, each containing a contiguous set of Transformer layers. A one‑time warm‑up profiling characterizes each block’s latency and memory footprint under different chunk sizes. This profile is used to determine the optimal partitioning and graph combination for the device’s NPU configuration.
\systemname~also maintains a token database that includes chat records and relevant corpus to facilitate draft generation.

\noindent $\bullet$ \textit{Online execution.}
When an augmented user request arrives, \systemname~performs the following stages:

\noindent \textit{Prefill}. 
The request is partitioned into fixed‑size chunks following the approach of prior mobile LLM systems~\cite{work:mllm, work:heterollm}. NPU executes these chunks using a long‑sequence optimized (prefill) graph.
\textit{Progressive graph scheduling} (§\ref{sec:tech1}). 
During prefill, the system gradually transitions to the decoding‑optimized graph through block-by-block switching, overlapping I/O costs with prefill computation (§\ref{sec:tech1}).
\textit{Distribution calibration}. 
After prefill, the system collects the output logits and applies in‑context distribution calibration (§\ref{sec:tech2}) to augment the draft database. This procedure aims to reduce rejection-caused recomputation by aligning drafts' token statistics with the model’s conditional distribution.
\textit{Speculative decoding}. 
With the draft database, \systemname~performs speculative decoding using the decoding‑optimized graph. It iteratively constructs tree-structured drafts and verifies them via a confidence-based policy that identifies and reuses valuable tokens (§\ref{sec:tech3}). When an end‑of‑sequence token is verified, \systemname~returns the merged output and commits the interaction into the database to facilitate future generation.


\subsection{Progressive Graph Scheduling}
\label{sec:tech1}

\noindent \textbf{Challenges of fixed-shape compute graphs.}
As discussed in §\ref{sec:motivation_sub1}, mobile NPUs require static compute graphs, forcing prefill and decoding to share the same graph despite their distinct input patterns: prefill processes long sequences (e.g., input length 256), while decoding handles short inputs (e.g., input length 1). Using a shared graph thus leads to inefficiency, while keeping two graphs resident in memory is wasting limited RAM budget. To address this, we consider switching between two specialized graphs: $G^1$ optimized for prefill and $G^2$ optimized for decoding. We denote the computation latency of a forward pass of model $\operatorname{P}$ on graph $G^x$ ($x \in \{1,2\}$) as $T_{\text{compute}}(\operatorname{P}; G^x)$.

\begin{figure}[!t]
  \centering
  \begin{subfigure}[b]{\linewidth}
    \centering
    \includegraphics[width=\linewidth]{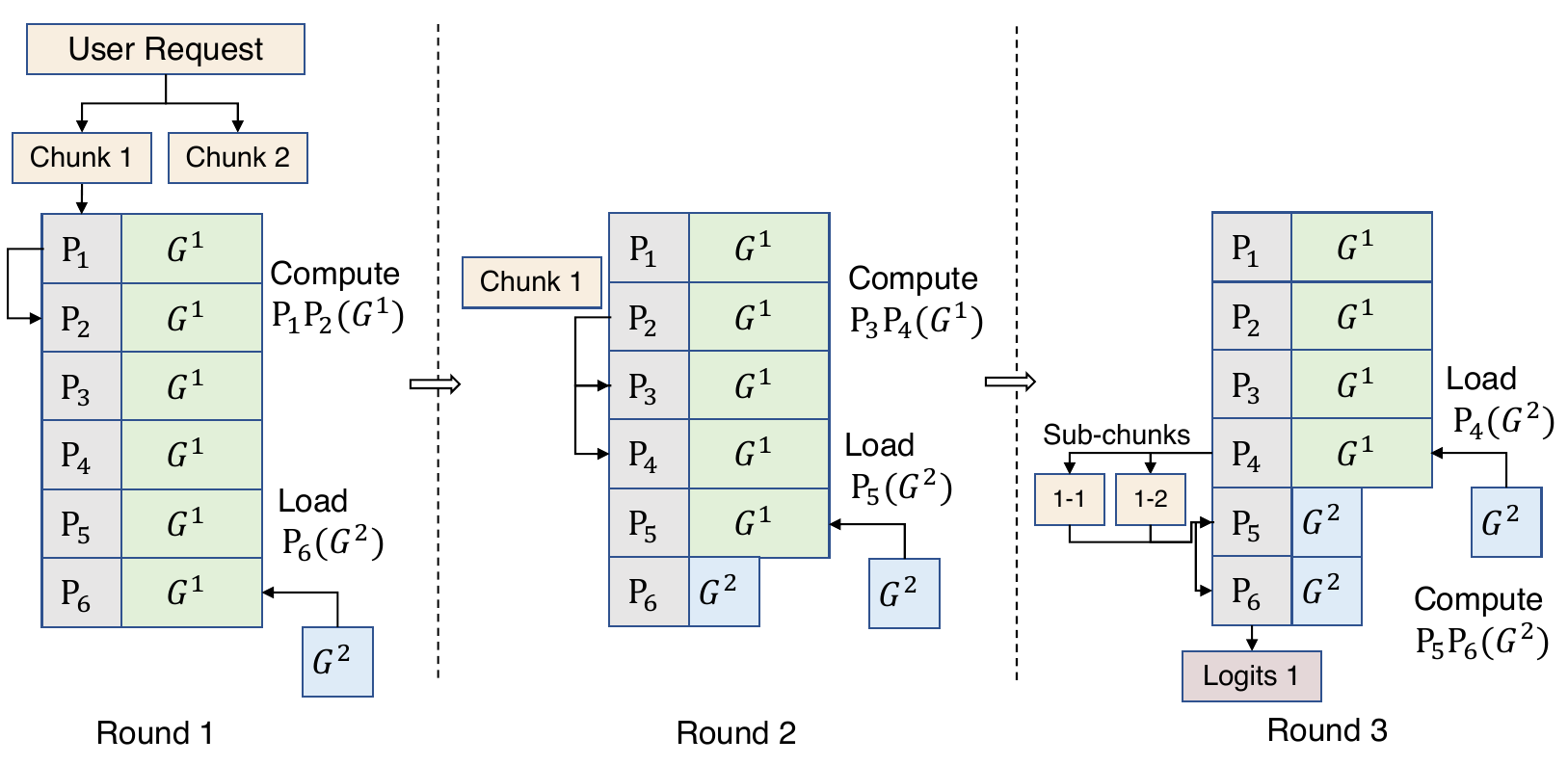}
    \caption{Progressive switching overlaps $G^2$ loading with computation across blocks.}
    \label{fig:t1_overview}
  \end{subfigure}
  \vspace{0.1cm}

  \begin{subfigure}[b]{\linewidth}
    \centering
    \includegraphics[width=\linewidth]{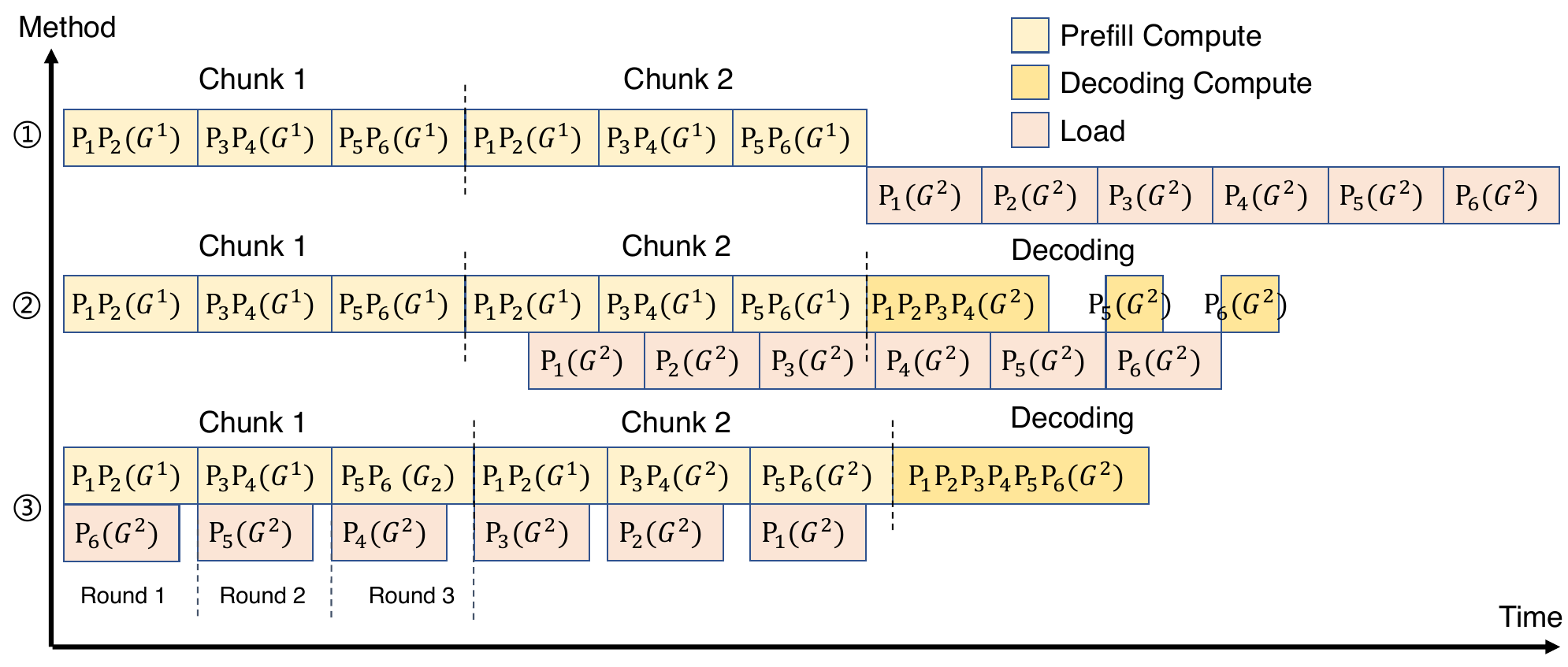}
    \caption{\systemname~(\textcircled{\scriptsize 3}) hides load latency compared to synchronous (\textcircled{\scriptsize 1}) and naive asynchronous (\textcircled{\scriptsize 2}) scheduling.}
    \label{fig:t1_demo}
  \end{subfigure}

    \caption{Progressive graph scheduling example with $N=6$ and two chunks. (a) workflow of computing chunk 1; (b) pipelines comparison under different scheduling. }
  \label{fig:t1}
  \vspace{-4pt}
\end{figure}

A naive solution is dynamically loading the decoding graph after prefill phase completes (Figure~\ref{fig:t1_demo}\textcircled{\scriptsize 1}). This \emph{synchronous loading} approach is inefficient as it delays decoding by I/O cost $T_{\text{load}}(\operatorname{P}; G^2)$.
To address this, an intuitive idea is \emph{asynchronous loading}, \modified{where the monolithic model $\operatorname{P}$ is partitioned into $N$ blocks $\{\operatorname{P}_1,\dots,\operatorname{P}_N\}$, each with equal amount of Transformer layers, and $G^2$ is loaded block-by-block after computing finishes} (Figure~\ref{fig:t1_demo}\textcircled{\scriptsize 2}). Yet this is suboptimal because: (i) following the existing approaches (§~\ref{sec:motivation}), the prefill input requires to be chunked for variable-length requests~\cite{work:mllm, work:heterollm}, so $G^1$'s switching is deferred to the last chunk; and (ii) load time $T_{\text{load}}(\operatorname{P}_i; G^2)$ may exceed compute time $T_{\text{compute}}(\operatorname{P}_j; G^x)$, leaving no effective overlap. 

\noindent \textbf{Progressive switching.} 
Instead, \systemname~amortizes the heavy switching cost across the entire prefill through two insights:

\noindent $\bullet$ \textit{Graph equivalence:} executing $G^2$ multiple times on shorter inputs is mathematically equivalent to a single execution of $G^1$ on longer inputs, allowing early switching. As \systemname~maximizes draft lengths (§\ref{sec:tech3}), $G^2$ will not be too short (e.g., chunk/4 of $G^1$), avoiding performance degradation.

\noindent $\bullet$ \textit{Chunk decomposition:} through fine-grained block partitioning and scheduling, the I/O payload of a single block can be masked by several chunked prefill iterations.

Figure~\ref{fig:t1} illustrates progressive scheduling with $N=6$ and two input chunks. We suppose that \systemname~decides to overlap loading a block by two blocks of computation based on the profiling results. Initially, all blocks are loaded with $G^1$. \systemname~computes $\operatorname{P}_1,\operatorname{P}_2$ with $G^1$ while loading $G^2$ for $\operatorname{P}_6$. Then $\operatorname{P}_3,\operatorname{P}_4$ execute while $G^2$ loads for $\operatorname{P}_5$. To compute $\operatorname{P}_5,\operatorname{P}_6$ under $G^2$, chunk 1 is split into sub-chunks matching $G^2$'s shape, with $G^2$ executed repeatedly. Prefill of chunk 1 finishes with output logits while half of the model is switched. The same procedure applies to chunk 2. In contrast, synchronous loading waits for all compute to finish, and naive asynchronous loading only overlaps in chunk 2, both incurring higher latency, as shown in Figure~\ref{fig:t1_demo}.

\noindent\textbf{Problem Formulation.}  
Formally, \modified{given $N$ model blocks (decided by offline profiling) that sequentially execute prefill for $C$ chunks}, \systemname~aims to determine the time to load each block, i.e., choosing blocks $\{\operatorname{P_i,\dots}\}$ to overlap load overhead of $\operatorname{P_j}$ by their computations at each compute-load round. The objective is to minimize the overall latency, which is equal to the maximum of the prefill completion time and the switch completion time. 

\noindent \textbf{Scheduling methodology.} This scheduling problem is NP-hard as it is essentially equivalent to the \emph{2-Partition} problem~\cite{work:partition}.
\modified{\systemname~addresses this via a greedy algorithm, choosing the least $k$ consecutive blocks $\{\operatorname{P}_{i},\dots,\operatorname{P}_{i+k-1}\}$ that overlap a block load to execute. This strategy enables fully masking the load overhead by computation to minimize switching overhead.}
\systemname~finds the smallest $k$ by estimating $\widehat{T}_{\text{load}}(\operatorname{P}_j; G^2)$ and $\widehat{T}_{\text{compute}}(\operatorname{P}_i; G^x_i)$ based on the profiling results.
To hide load latency, the sum of the next $k$ blocks’ compute time must be larger than the load overhead of the last block not switched, formulated as:
\begin{align}
\widehat{T}_{\text{load}}(\operatorname{P}_j; G^2) 
\;\le\; \sum_{\ell=0}^{k-1}\widehat{T}_{\text{compute}}(\operatorname{P}_{i+\ell}; G^x_{i+\ell}).
\label{eq:bound}
\end{align}
\systemname~greedily chooses the minimal feasible $k$ that satisfies Equation~\ref{eq:bound}:
\begin{align}
k^\star \;=\;
\min\Big\{ k\ge 1 \;\big|\;
\widehat{T}_{\text{load}}(\operatorname{P}_j; G^2) \le \sum_{\ell=0}^{k-1}\widehat{T}_{\text{compute}}(\operatorname{P}_{i+\ell}; G^x_{i+\ell})
\Big\}.
\label{eq:k}
\end{align}

Then \systemname~executes $\operatorname{P}_{i},\dots,\operatorname{P}_{i+k^\star-1}$ under current graphs while loading $G^2$ for $\operatorname{P}_j$. This process repeats until all blocks are switched.

\subsection{In-Context Distribution Calibration}
\label{sec:tech2}

\begin{figure}[!t]
  \centering
  \includegraphics[width=\linewidth]{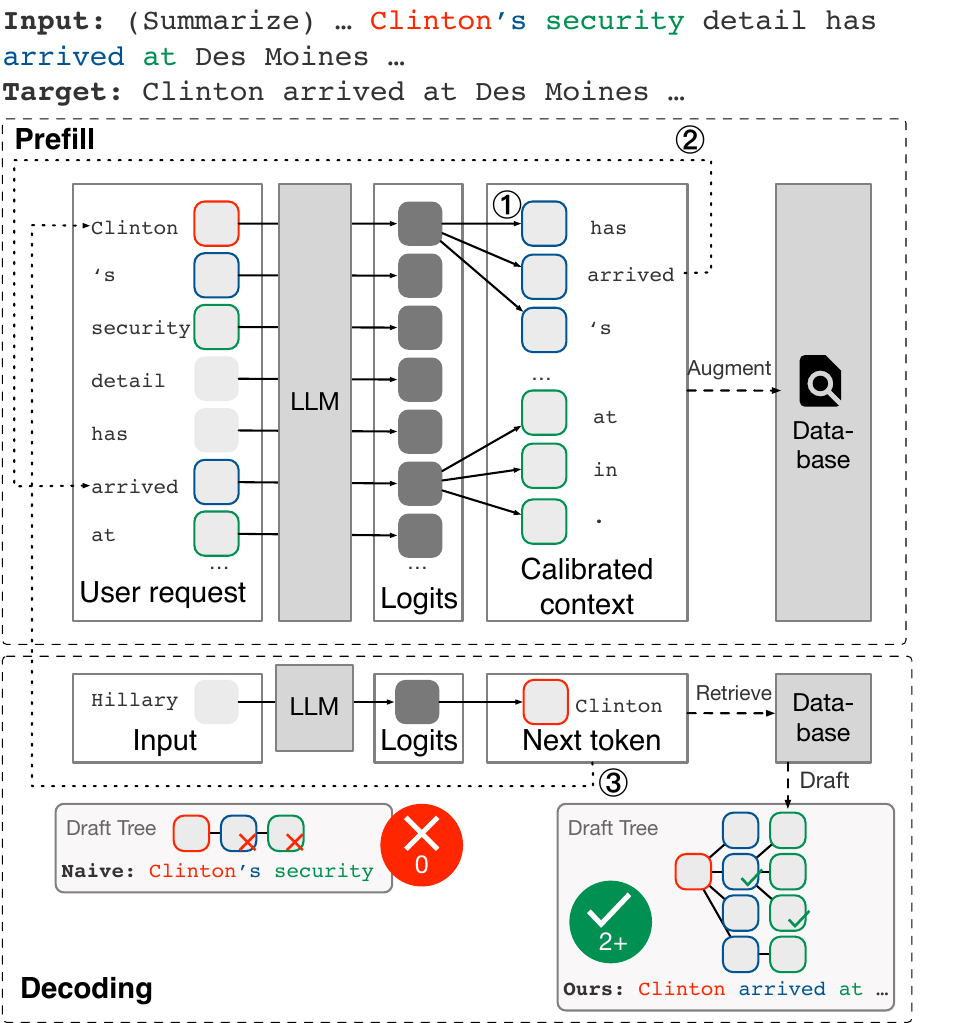}
  \caption{Overview of in-context distribution calibration with a summarization example. }
  \label{fig:t2_overview}
  \vspace{-4pt}
\end{figure}

\noindent \textbf{Wasted computation due to rejection.}
\modified{To achieve lossless acceleration, speculative decoding rejects all draft tokens after the first incorrect token. Each rejection forces recomputation and falling back to token-by-token decoding, neutralizing speculative parallelism and wasting NPU cycles. This verification strategy requires lexical identity, resulting that semantically valid drafts are rejected.
This lexical mismatches frequently arise in mobile RAG workloads such as summarization and QA, where retrieved text originates from human-written content with phrasing styles that deviate from the model’s learned patterns.}
As shown in Figure~\ref{fig:t2_overview}, given a draft “\textit{Clinton’s security detail has arrived at Des Moines},” the model’s own distribution predicts “\textit{Clinton arrived at Des Moines}.” Although the semantics align, lexical mismatch at the token level (e.g., \textit{'s} vs.\ \textit{arrived}) triggers rejection~\cite{work:pld, work:rest, work:sam}. The verification then produces only one accepted token (\textit{arrived}), forcing recomputation for the rest (\textit{security detail}). Across NPU pipelines, this leads to redundant graph executions and degraded decoding throughput.

\noindent \textbf{Design goal.}
We aim to reduce wasted NPU computation caused by draft rejection, without compromising semantic accuracy. The key idea is to align drafts with the model’s intrinsic distribution such that they better match what the model is likely to verify.
A straightforward solution is to precompute a calibrated database by running large volumes of user inputs through the LLM, which suffers from poor generalization to unseen queries and requires prohibitive storage and offline processing. Another method is to fine-tune the LLM to adapt to specific scenarios, which also consumes lots of resources and is hard to cover the wide diversity of RAG tasks.


\noindent \textbf{Leverage prefill logits.}
During prefill, NPUs already compute next-token logits for all context tokens as part of the model’s forward pass. These logits encode the LLM’s internal probability distribution and naturally reveal how the model interprets the current task context. By reusing these prefill logits, we can estimate the model’s local token transition space and calibrate the retrieval distribution without additional forward passes or fine-tuning. 
We call this mechanism \textit{in-context distribution calibration}, an online, lightweight technique that exploits prefill logits to dynamically align drafts with model-side token probabilities, effectively avoiding wasteful computation on rejected drafts.

\noindent \textbf{Workflow.}
After prefill, \systemname~first collects predicted logits for all positions. Then \systemname~builds a token-level calibration structure that approximates the model’s local generation distribution by 2-step depth-first search:

\noindent $\bullet$ \textit{Sample} (Figure~\ref{fig:t2_overview}~\textcircled{\scriptsize 1}). For each token in the user request (e.g., “\textit{Clinton}”), \systemname~samples top-probability successors (\textit{has}, \textit{arrived}, \textit{'s}) from its prefill logits. Each \textless context, prediction\textgreater\ pair forms a calibrated anchor (e.g., \textit{Clinton–arrived}) that reflects the model’s token-level preferences.

\noindent $\bullet$ \textit{Search} (Figure~\ref{fig:t2_overview}~\textcircled{\scriptsize 2}).  \systemname~scans the retrieved context to locate occurrences of these model-preferred anchors. Once matched (e.g., “\textit{arrived}”), it recursively expands successors using that token’s prefill logits (e.g., \textit{at}, \textit{in}, or \textit{.}). The result is a token tree representing the model-aligned transition structure.

The calibrated token tree is then cached as a lightweight source for future drafting (Figure~\ref{fig:t2_overview}~\textcircled{\scriptsize 3}). Instead of fetching raw phrases, drafts are now selected from tokens that the model itself deems plausible given the current RAG context. This alignment reduces the likelihood of verifier rejection and thus lowers recomputation frequency.
In the running example, “\textit{Clinton}” yields a calibrated draft “\textit{Clinton arrived at},” which matches the model’s predicted trajectory and passes verification without additional recomputation. The overhead discussion is detailed in §~\ref{sec:memory_overhead} and §~\ref{sec:latency_breakdown}.


\subsection{NPU-Optimized Draft Extension}
\label{sec:tech3}

\noindent \textbf{Short drafts underutilize NPU.}
\modified{NPUs are optimized for dense, long-running workloads that amortize weight loading and maximize matrix reuse across many tokens. However, existing workloads are typically short: most drafts contain 2-8 tokens, well below the length needed to saturate systolic arrays, as shown in Figure~\ref{fig:cpu_npu_decoding_speed}.
Simply lengthening drafts is ineffective because irrelevant tokens are immediately rejected, forcing recomputation and wasting NPU cycles. Therefore, extending drafts with plausible tokens is crucial for efficient NPU execution. }

\noindent \textbf{Observation: rejected tokens are not always incorrect.}
While SD rejects drafts with lexical mismatches, these tokens often encode semantically correct continuations. For example, when summarizing “\textit{Clinton’s security detail has arrived at Des Moines},” the model may reject “\textit{security}” due to token-level mismatch, yet the phrase “\textit{Clinton arrived at}” appears again in subsequent verification. Thus, the earlier rejected segment partially overlaps with future correct output, representing “wasted” computation that could be reused to extend future drafts and reduce underutilized NPU cycles.

\begin{table}[tb]
\footnotesize
  \begin{tabular}{lrrr}
    \toprule
    Case & Pattern & Reusable & Ratio  \\
    \midrule
    \textbf{Missing} & $z_e\dots z_{e+\delta}x_{e}\dots x_{e+\epsilon}$ & $x_e\dots x_{e+\epsilon}$ & 15.4\% \\
    \textbf{Synonym} & $z_e\dots z_{e+\delta}x_{e+\gamma}\dots x_{e+\epsilon}$ & $x_{e+\gamma}\dots x_{e+\epsilon}$ & 23.1\% \\
    \textbf{Redundant} & $z_e\dots z_n$ & $\emptyset$ & 61.5\% \\
    \bottomrule
  \end{tabular}
  \caption{Given draft $T,x_1\dots x_n$ which misaligns with target $z_1\dots z_{n+1}$ at position $e$, the possible patterns of segment $z_e\dots z_n$ and corresponding reusable tokens. In cases \textit{missing} and \textit{synonymous}, tokens after $x_{e+\epsilon}$ are omitted for simplicity.
  }
  \label{tab:cases}
  
\vspace{-4pt}
\end{table}

\noindent \textbf{Empirical analysis of reuse opportunities.}
We analyze rejection causes by comparing drafts $\mathbf{x}=[T,x_1,\dots,x_n]$ with model predictions $\mathbf{y}=[y_1,\dots,y_{n+1}]$ and expected sequences $\mathbf{z}=[z_1,\dots,z_{n+1}]$.
Suppose a rejection occurs at position $e$ where $x_e \neq y_e$. The accepted prefix is $x_1,\dots,x_{e-1},y_e$.
Our manual annotation across a summarization dataset (Table~\ref{tab:cases}) reveals that nearly 40\% of rejected tokens appear again within subsequent decoding steps, indicating that they can be safely reused to extend NPU workloads.
Specifically, we categorizes three major rejection patterns in Table~\ref{tab:cases}:

\noindent $\bullet$ \textit{Missing:} The expected output includes additional tokens absent from the draft, but subsequent draft tokens remain valid.
Reusing $x_e\dots x_{e+\epsilon}$ preserves the valid suffix.

\noindent $\bullet$ \textit{Synonym:} The rejected tokens are semantically equivalent but lexically different (e.g., “\textit{car}” vs.\ “\textit{vehicle}”).
Retaining the following segment $x_{e+\gamma}\dots x_{e+\epsilon}$ allows reusing once the synonym mismatch resolves.

\noindent $\bullet$ \textit{Redundant:} The entire segment diverges from the target sequence and must be discarded.

\noindent \textbf{Reuse strategy.}
\systemname~dynamically identifies reusable token segments from previously rejected drafts and appends them to new drafts to form NPU-friendly workloads.
Formally, \systemname~aims to: (1) determine the start and end positions $\gamma, \epsilon$ of the maximum reusable segment; (2) estimate the reuse lifetime $\delta$, i.e., how many iterations the reused segment should be reserved for verification.

Enumerating all possible reuse strategies is of $O(N^3)$ complexity, which is not suitable for online scheduling.
Therefore, \systemname~adopts a confidence-based and length-first reuse strategy, assuming that drafts aligned with model predictions are more likely to be accepted in future steps. Therefore, we identify and retain the longest segment of $\mathbf{x}$ such that $x_i = y_i$ for all $i \in [e+\gamma, e+\epsilon]$, by solving:
\begin{align}
    \gamma,\epsilon=\underset{\gamma,\epsilon}{\text{argmax}}(\epsilon-\gamma)  \quad \text{s.t. } y_i = x_i,\ \forall\ i \in [e+\gamma, e+\epsilon].
    \label{eq:reuse}
\end{align}
This reused segment is retained across at least one decoding step and discarded once its accumulated lifetime or total draft length exceeds a predefined threshold.
Thus, \systemname~effectively reuse wasted tokens and reduce underutilized ones, saving 70.5\% of token computation without sacrificing accuracy.
The additional overhead is discussed in §~\ref{sec:latency_breakdown}.

 \section{Evaluation} \label{sec:evaluation}

\begin{figure*}[tbp]
  \centering
  \includegraphics[width=1.0\textwidth]{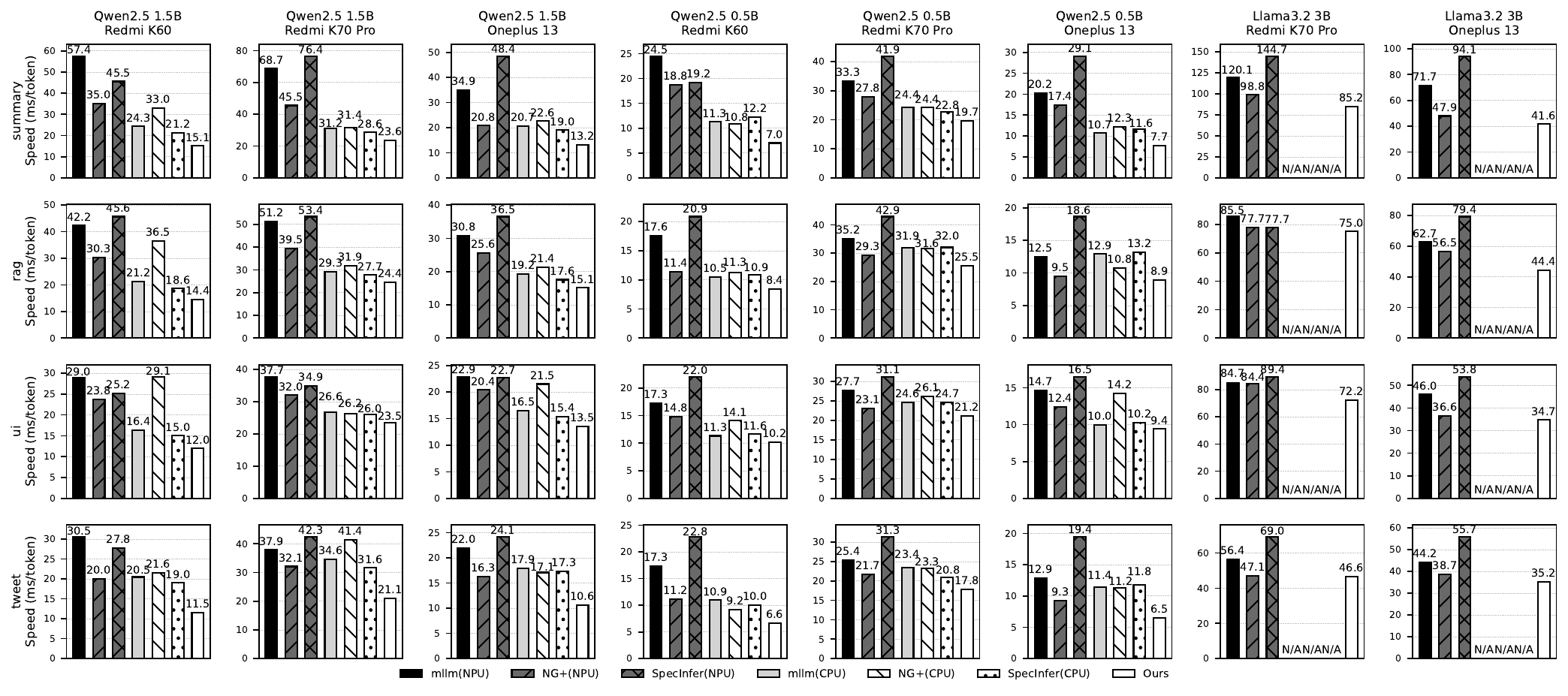}
  \caption{Per-token latency (ms/token) comparison on different datasets, devices and models. }
  \label{fig:speed_all}
  \vspace{-4pt}
\end{figure*}


\subsection{Experiment Setup}
\label{sec:setup}

\noindent\textbf{Hardware}.
We evaluate \systemname~on three representative smartphones with Qualcomm SoCs: \textit{Redmi K60 Pro} (Snapdragon 8 Gen 2, Android 13, 12\,GB RAM), \textit{Redmi K70 Pro} (Snapdragon 8 Gen 3, Android 14, 16\,GB RAM), and \textit{OnePlus 13} (Snapdragon 8 Gen Elite, Android 14, 24\,GB RAM).
All devices are evaluated under real mobile conditions with the CPU frequency governed by the Android OS's dynamic voltage and frequency scaling controller.


\noindent\textbf{Datasets}.
We evaluate four common mobile tasks collected from prior work: document summarization (\textit{summary})~\cite{dataset:mt_bench}, RAG-based question answering (\textit{rag})~\cite{dataset:mt_bench}, UI automation (\textit{ui})~\cite{dataset:ui} and automatic message reply (\textit{tweet})~\cite{dataset:lamp}.

\noindent\textbf{Models}.
We use Qwen2.5-0.5B-Instruct, Qwen2.5-1.5B-Instruct~\cite{qwen25} and LLaMA3.2-3B-Instruct~\cite{llama3}, exploring the effect of different model size.

\noindent\textbf{Baselines}.
\modified{
We compare with mllm and two on-device LLM inference systems that optimizes speculative decoding: SpecInfer~\cite{work:specinfer} for model-based SD and NG+~\cite{work:ng+} for retrieval-based SD. To eliminate performance gaps from backend design differences, we implement their algorithms atop mllm
\footnote{Other existing mobile LLM inference systems with NPU offloading support such as PowerInfer-v2~\cite{work:powerinfer2} and HeteroLLM~\cite{work:heterollm} are not open-source, making them unsuitable for direct implementation. \systemname, as a system-level optimization, is orthogonal and integrable with T-MAN~\cite{work:tman}, so we do not compare it directly.
}.
While all baselines use NPU for prefill, which is a widely-adopted approach for both industry and academic~\cite{work:mllm, work:heterollm, work:transformer_lite, work:powerinfer2}, we evaluate decoding baselines on different backends, i.e., CPU and NPU to augment our experiments. In summary, our baselines include}:

\noindent (1) \textit{mllm(NPU)}: use NPU for both prefill and decoding\footnote{mllm~\cite{work:mllm} and other on-device LLM engines ~\cite{work:heterollm,work:powerinfer2} offload precision-sensitive operations, such as attention, to CPU/GPU to improve response quality. \systemname~follows this pattern even in the NPU vanilla mode.};

\noindent (2) \textit{mllm(CPU)}: use NPU for prefill and CPU for decoding;

\noindent (3) \textit{NG+(NPU)}: apply retrieval-based SD from NG+~\cite{work:ng+} to mllm-NPU. We use SAM~\cite{work:sam} as retrieval algorithm, which is the SOTA method to draft from both request prompt and extra database;

\noindent (4) \textit{NG+(CPU)}: similar to (3) but decoding on CPU;

\noindent (5) \textit{SpecInfer(NPU)}: apply model-based SD from SpecInfer~\cite{work:specinfer} to mllm-NPU. We use EAGLE-2~\cite{work:eagle2} as draft model, which is the SOTA model-based SD approach that uses a pretrained transformer-like model for drafting;

\noindent (6) \textit{SpecInfer(CPU)}: similar to (5) but decoding on CPU.


\noindent\textbf{Hyper-parameter settings}. \modified{In practice, we observe that the $N=2$ partitioning is sufficient for our selected models and devices to fully overlap the graph switching costs by computation. Therefore, we fix the model partitioning schema of $N=2$ across all evaluation. For the retaining threshold in draft extension technique, we set a limit of 32, which is the least draft length to saturate NPU as shown in Figure~\ref{fig:cpu_npu_decoding_speed}}.

\noindent \textbf{Metrics and Configuration}.
We evaluate average per-token latency, energy consumption and peak memory. We exclude accuracy comparison as \systemname~follows the strict verify schema in SD~\cite{work:sd_is_lossless} and achieves lossless generation.
Energy is measured via Android’s virtual file system under \verb|/sys/class/power_supply| by profiling every 100ms. The Redmi K70 Pro is excluded from energy results as the device lacks root access. Experiments are repeated three times and we report the average numbers.

\subsection{Overall Performance}
\label{sec:result_analysis}

\noindent \textbf{Per-Token Latency.} \textbf{\systemname~is faster and achieves improvements ranging 1.06--3.81×}, as shown in Figure~\ref{fig:speed_all}.

\noindent $\bullet$ \textit{Compared to baselines without SD (mllm(NPU) and mllm(CPU))}, \systemname~consistently reduces per-token latency by 1.14--3.81× and 1.06--1.78×. These performance gains stem from both the NPU-optimized SD method and the specialized compute graphs.
Larger improvements are observed on small-to-medium models (e.g., Qwen2.5 0.5B and 1.5B) as their inference is less compute-intensive.
Comparing different datasets, we find that tasks with higher context similarity such as \textit{summary}, exhibit higher speed gains (1.41--3.80×, 1.24--1.61×) since SD is more effective.

\noindent $\bullet$ \textit{Compared to baselines with SD,} \systemname~consistently improves by 1.11--2.53× than NG+, 1.09--1.80× than SpecInfer. This arises because NG+ ignores the inefficiency of computing short drafts on NPUs while SpecInfer suffers from the latency of running another parametric model.
Improvements of applying SD on CPU are moderate (0.92--1.13× for NG+, 0.56--1.18× for SpecInfer), as LLM inference on mobile CPU is compute-bound due to the limited CPU capacity.

\begin{figure*}[tbp]
  \centering
  \includegraphics[width=1.0\textwidth]{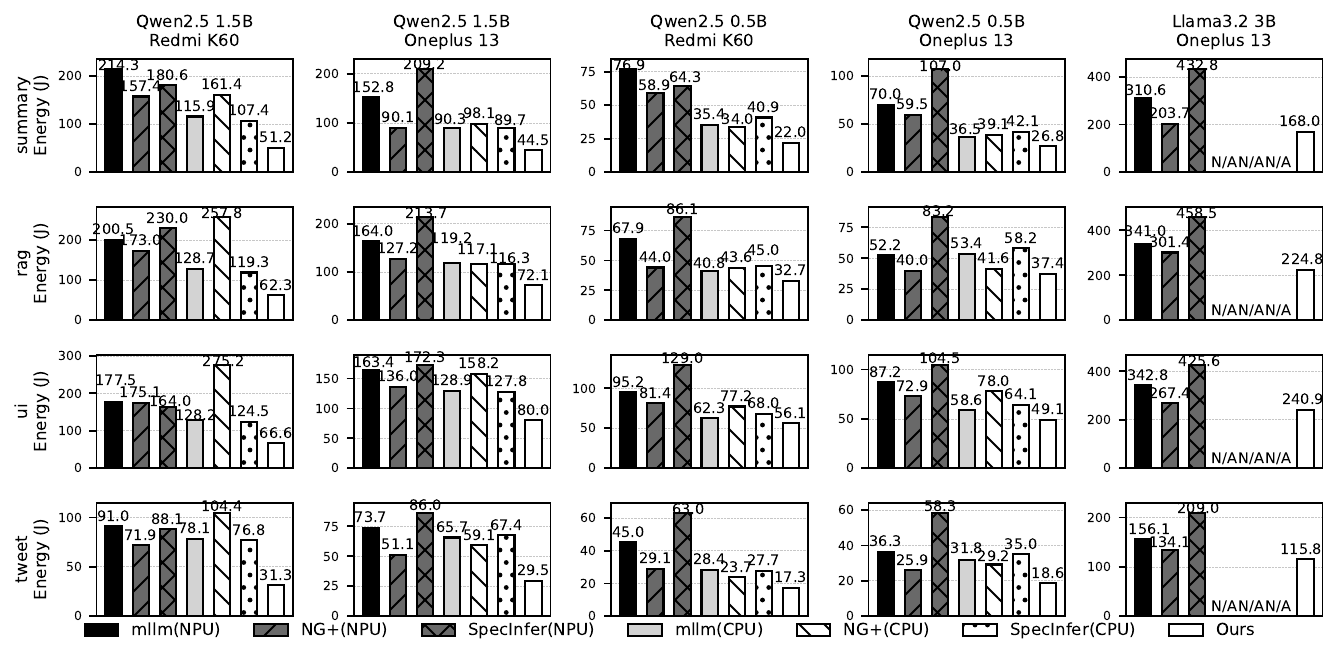}
  \caption{Energy consumption (J) comparison on different datasets,  devices and models. }
  \label{fig:energy_all}
\vspace{-4pt}
\end{figure*}

\noindent \textbf{Energy}. \textbf{\systemname~is more energy-efficient and reduces energy consumption by 1.07--4.71×} (Figure~\ref{fig:energy_all}).

\noindent $\bullet$ \textit{Compared to baselines without SD (mllm(NPU) and mllm(CPU))}, \systemname~significantly reduces energy consumption by 1.35--4.18× and 1.11--2.50×. The improvements stem jointly from the higher generation speed and the avoidance of CPU computation, allowing energy-efficient utilization of NPUs. On larger models (e.g., Llama3.2 3B), the effectiveness becomes less pronounced (1.42--1.85×) as relatively longer inference time results in the device continues working under a high power consumption.

\noindent $\bullet$ \textit{Compared to baselines with SD}, \systemname~consumes significantly less energy than NG+ (1.07--3.07×) and SpecInfer (1.77--4.71×). Energy saving of NG+ is limited due to frequent draft rejection. SpecInfer suffers from energy overhead amplified by drafter computation. For instance, for Qwen2.5 0.5B, \systemname~reduces energy 1.40--3.50× compared with mllm(NPU), while NG+ reduces energy by only 1.17--1.54× and SpecInfer even increases energy consumption. 

\noindent \textbf{Comparison Across Devices.}
We find that NPU computation on Redmi K70 Pro is much slower than on other devices, while Redmi K60 Pro and Oneplus 13 demonstrate relatively similar performance patterns. This results in less profound latency improvements on Redmi K70 Pro (e.g., 1.14--2.91× compared to mllm(NPU) and 1.13--1.64× compared to mllm(CPU)) as compute becomes the bottleneck. In terms of energy consumption, Oneplus 13 consumes more energy than Redmi K60 Pro, e.g., average 53.85\,J and 56.52\,J per request for Qwen2.5 1.5B.

\subsection{Memory Overhead}
\label{sec:memory_overhead}

\begin{figure}[tbp]
  \centering
  \includegraphics[width=\linewidth]{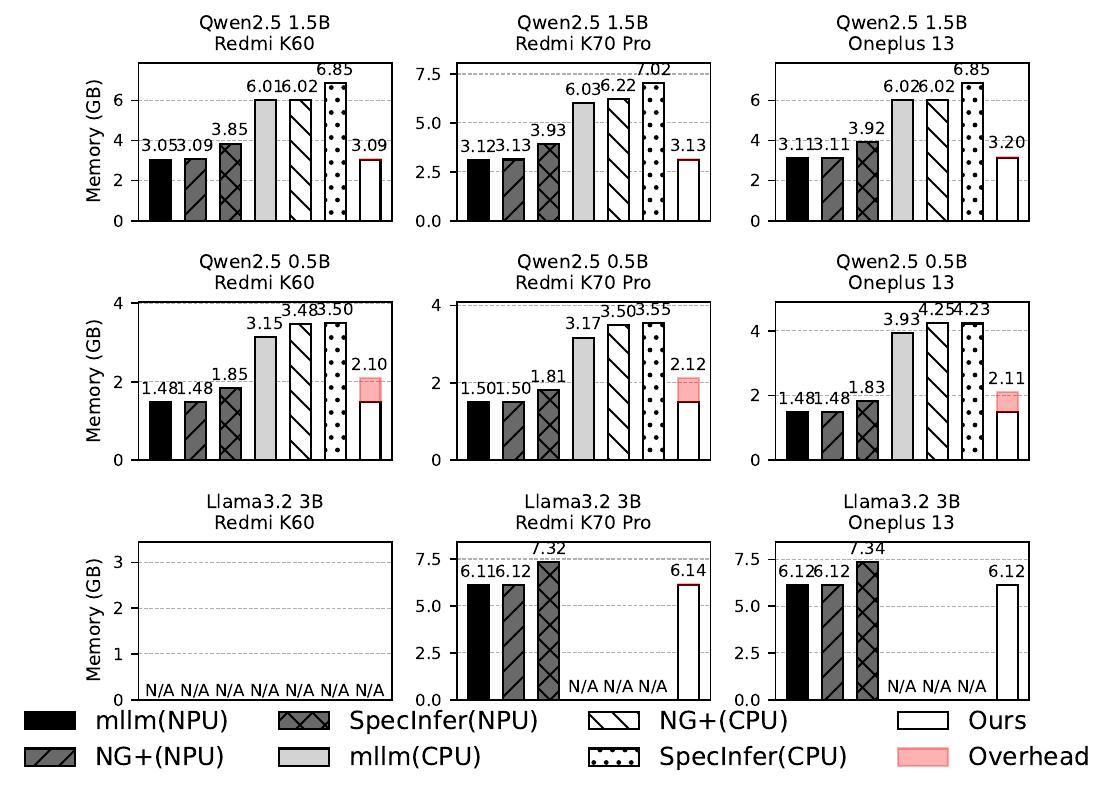}
  \caption{Peak memory comparison (GB) on different devices and models. }
  \label{fig:memory_all}
  \vspace{-4pt}
\end{figure}

As shown in Figure~\ref{fig:memory_all}, \systemname~maintains a memory footprint close to mllm(NPU) with an additional overhead less than 500\,MB across all evaluations. 
For example, \systemname~requires 3.14\,GB for Qwen2.5 1.5B, nearly identical to mllm(NPU) (3.05--3.12\,GB), while mllm(CPU) requires over 6 GB. This gap comes from maintaining separate weights for CPU and NPU in heterogeneous baselines.

The memory overhead of \systemname~primarily arises from distribution calibration and context-relevant database structures, which is manageable for mobile devices with limited memory.
In contrast, SpecInfer's memory usage increases noticeably (0.84--1.22\,GB) due to maintaining the drafter model, highlighting that \systemname~delivers superior efficiency without sacrificing memory scalability.


\subsection{Latency Breakdown}
\label{sec:latency_breakdown}

To analyze the runtime characteristics and latency overhead of \systemname, we decompose the end-to-end inference latency into prefill, decoding and potential overhead. Measurements are conducted on Oneplus 13 under the \textit{summary} benchmark.

\begin{figure}[tbp]
  \centering
  \includegraphics[width=\linewidth]{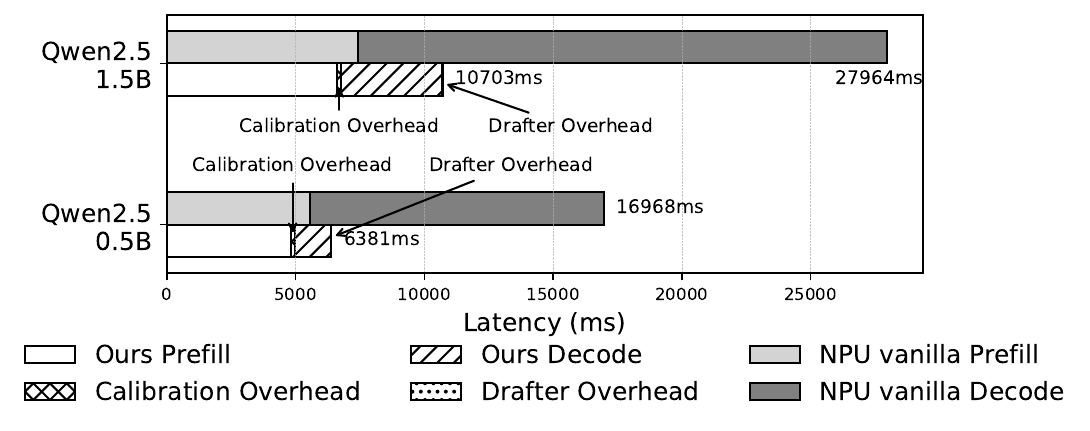}
  \caption{Breakdown of inference latency (ms). }
  \label{fig:latency_breakdown}
  \vspace{-4pt}
\end{figure}

The results (Figure~\ref{fig:latency_breakdown}) show that \textbf{\systemname~significantly accelerates both prefill and decoding compared to mllm(NPU) baseline, with little additional overhead}. Specifically, decoding of mllm(NPU) consumes a substantial portion of total latency (67--74\%), highlighting the significance of introducing specialized optimization. \systemname~directly addresses this by integrating NPU-optimized speculative decoding, reducing decoding latency by 5.25--8.15×.

The prefill overhead is due to sampling tokens from logits for distribution calibration, which costs about 2\,ms for each input token. The decoding overhead is caused by draft tree construction. Benefited from the low time complexity of suffix automaton, our drafter costs less than 3\,ms for each decoding step. Overall, our lightweight techniques incur negligible overhead compared to the overall inference cost. 


\subsection{Ablation Study}
\label{sec:ablation}

\noindent \textbf{Effectiveness of three techniques.}
Experiments are conducted on Redmi K60 Pro using the \textit{summary} benchmark across four baselines: (1) \textit{mllm(NPU)}, (2) \textit{NPU-SD} (\textit{+SD}), (3) \textit{NPU-Graph} (\textit{+G}), and (4) \textit{NPU-SD-Graph} (\textit{+G+SD}). Baselines (2)(4) use SAM~\cite{work:sam} without our optimization, while (3)(4) integrate \textit{progressive graph scheduling} into (1) and (2). Results are reported in Table~\ref{tab:ablation_graph}.  

\begin{table}[t]
\footnotesize
  \begin{tabular}{lrrrrr}
    \toprule
    \textbf{Model} & \textbf{NPU} & \textbf{+G} & \textbf{+SD} & \textbf{+G+SD} & \textbf{Ours}\\
    \midrule
    \textbf{Qwen2.5-1.5B} & 57.41 & 47.72 & 35.04 & 25.35 & 15.08 \\
    \textbf{Qwen2.5-0.5B} & 24.46 & 23.43 & 18.75 & 18.45 & 7.17 \\
    \textbf{LLaMA3.2-3B} & 71.71 & 70.03 & 47.93 & 46.25 & 41.62\\
    \bottomrule
  \end{tabular}
  \caption{Ablation on average per-token latency (ms/token). }
  \label{tab:ablation_graph}
\vspace{-4pt}
\end{table}

We observe that progressive graph scheduling provides consistent speed improvements regardless of whether SD is applied. Note that the performance gains are amortized across the end-to-end generation, thus its improvement is less pronounced for larger LLM whose computation is the main bottleneck, or small model that requires relatively short time to switch the graph compared to the whole generation latency. 
The results also show that our optimized SD contributes to significant performance gains across three models, reducing latency from 25.35\,ms to 15.08\,ms for Qwen2.5-1.5B.

 
\noindent \textbf{Effectiveness in accept length.}
To analyze the effectiveness of SD optimization, we evaluate the accept length gains applying our techniques to existing methods (PLD~\cite{work:pld}, SAM~\cite{work:sam}). PLD drafts from current context (i.e., user request) only while SAM drafts from both current context and an extra database. Experiments are conducted on a cloud server equipped with A100 GPUs. The results in Table~\ref{tab:ablation_sd} show consistent improvements across benchmarks and models. Our techniques improve the accept length of PLD by 1.05--1.25×, improve SAM by 1.06--1.24×. These improvements directly translate to higher NPU efficiency, as discussed in §~\ref{sec:motivation_sub2}. 
These results also demonstrate that our techniques provide orthogonal enhancements with existing SD methods, making it effortless to integrate advanced techniques.  

\begin{table}[t]
\centering
\footnotesize
\begin{tabular}{lllll}
\toprule
\textbf{Dataset} & \textbf{Method} & \textbf{Q1.5B} & \textbf{Q7B} & \textbf{L1B} \\
\midrule
\multirow{4}{*}{\textbf{summary}} 
& PLD  & 2.53 & 1.55 & 1.42 \\
& Ours & 2.91 {\scriptsize(1.15×)} & 1.84 {\scriptsize(1.18×)} & 1.70 {\scriptsize(1.19×)} \\
& SAM  & 3.60 & 1.96 & 1.68 \\
& Ours & 4.21 {\scriptsize(1.17×)} & 2.43 {\scriptsize(1.24×)} & 2.07 {\scriptsize(1.24×)} \\
\midrule

\multirow{4}{*}{\textbf{rag}} 
& PLD  & 3.19 & 1.63 & 1.77 \\
& Ours & 3.73 {\scriptsize(1.17×)} & 1.89 {\scriptsize(1.16×)} & 2.11 {\scriptsize(1.19×)} \\
& SAM  & 4.42 & 2.29 & 2.25 \\
& Ours & 5.47 {\scriptsize(1.24×)} & 2.80 {\scriptsize(1.22×)} & 2.66 {\scriptsize(1.19×)} \\
\midrule

\multirow{4}{*}{\textbf{ui}} 
& PLD  & 2.28 & 1.64 & 1.57 \\
& Ours & 2.64 {\scriptsize(1.16×)} & 1.73 {\scriptsize(1.05×)} & 1.82 {\scriptsize(1.16×)} \\
& SAM  & 3.56 & 2.44 & 1.72 \\
& Ours & 4.01 {\scriptsize(1.13×)} & 2.59 {\scriptsize(1.06×)} & 2.05 {\scriptsize(1.20×)} \\
\midrule

\multirow{4}{*}{\textbf{tweet}} 
& PLD  & 2.53 & 2.06 & 1.49 \\
& Ours & 3.17 {\scriptsize(1.25×)} & 2.52 {\scriptsize(1.22×)} & 1.72 {\scriptsize(1.15×)} \\
& SAM  & 3.57 & 2.98 & 1.79 \\
& Ours & 3.82 {\scriptsize(1.07×)} & 3.70 {\scriptsize(1.24×)} & 2.18 {\scriptsize(1.22×)} \\
\bottomrule
\end{tabular}

\begin{flushleft}
\footnotesize{Q1.5B = Qwen2.5 1.5B, Q7B = Qwen2.5 7B, L1B = LLaMA3.2 1B.}
\end{flushleft}
\caption{Ablation on accept length.}
\label{tab:ablation_sd}

\vspace{-4pt}
\end{table}

\section{Related Work} \label{sec:related_work}


\noindent \textbf{LLM inference on mobile NPU}. 
Mobile SoCs are increasingly equipped with high-performance NPUs~\cite{trend:oppo, trend:samsung, trend:huawei}, offering new opportunities for LLM acceleration~\cite{work:powerinfer2, work:heterollm, work:bluelm, work:mllm, work:agent_xpu}. mllm~\cite{work:mllm} and PowerInfer-V2~\cite{work:powerinfer2} offload precision-insensitive operations to NPU, while HeteroLLM~\cite{work:heterollm} pipelines GPU and NPU. Test-time-compute NPU~\cite{work:ttc} improves response quality by NPU-aware quantization. T-MAN~\cite{work:tman} reduces quantization overheads by kernel refinement. \systemname~complements these work by incorporating system-algorithm co-design to optimize RAG inference.


\noindent \textbf{Retrieval-augmented generation} enhances LLM's knowledge using relevant documents from an external database, enabling effective context-aware generation~\cite{work:rag}. Recent work explores advanced database architectures such as knowledge graph~\cite{work:graph_rag1, work:graph_rag2, work:graph_rag3}. Other efforts optimize retrieval with adaptive strategies~\cite{work:adaptive_rag, work:edge_rag, work:webanns} and improve memory management for long contexts~\cite{work:infinigen, work:page_attention}.

\noindent \textbf{Speculative decoding} uses LLM as a verifier for parallel computing~\cite{work:early_speculative_decoding, work:speculative_sampling, work:speculative_decoding}. Model-based SD uses an auxiliary parametric model for drafting, typically a smaller LLM~\cite{work:eagle, work:eagle2, work:medusa, work:glide, work:eagle_3}. Although these approaches improve the accept ratio, they introduce significant drafting overhead~\cite{work:sam}. Retrieval-based SD avoids this cost by using lightweight retrievers for drafting in the cost of low accept ratios~\cite{work:pld, work:pld+, work:sam, work:rapid, work:chunk_sd, work:dresd}. 
\systemname~advances them through hardware–algorithm co-design, integrating it with mobile NPUs for efficient acceleration.

\section{Conclusion}
\label{sec:conclusion}
We present \systemname, the first on-device RAG system enabling efficient end-to-end NPU offloading. \systemname~employs a system-algorithm co-design: it introduces progressive graph scheduling to mask graph switching latency, and novel speculative decoding optimizations (distribution calibration and draft extension) to saturate NPU utilization. Extensive evaluation on commercial smartphones demonstrates that \systemname~outperforms state-of-the-art systems with 1.06$\times$--3.81$\times$ speedup and 1.07$\times$--4.71$\times$ energy reduction.




\bibliographystyle{ACM-Reference-Format}
\bibliography{refs}

\appendix
\section{NP-hardness Proof of the Graph Scheduling Problem}
\label{appendix:proof}

We prove that the graph scheduling problem described in Section~\ref{sec:tech1} is NP-hard via a reduction from the \textit{2-Partition} problem~\cite{work:partition}.

\noindent \textbf{Graph scheduling problem}.
Given $N$ model blocks that sequentially execute prefill for $C$ chunks, each block $i$ must execute a prefill for every chunk and perform exactly one graph switch from $G^1$ to $G^2$. Using $G^2$ to prefill is slower than using $G^1$. Switches can run in parallel with the computation of other blocks, while switches / computation of different blocks cannot overlap with each other.
We aim to determine the time to switch each block with an objective to minimize the maximum of the prefill completion time and the switch completion time. 


\noindent \textbf{2-Partition problem}.
Given a set of positive integers $a_1,\dots,a_m$ with sum $S=2B$, determine whether there exists a subset $I\subseteq\{1,\dots,m\}$ such that $\sum_{j\in I} a_j = B$.

\noindent\textbf{Construction of the scheduling instance.} Given a Partition instance $a_1,\dots,a_m$, we construct a scheduling instance as follows:
\begin{itemize}[leftmargin=15pt]
    \item Let $N = m+1$. Indices $1, \dots, m$ correspond to the $m$ items of the Partition instance, and the block $0$ (or equivalently $m+1$) serves as a long block.
    \item Let the number of chunks be $C=2$.
    \item For each $j = 1, \dots, m$ (corresponding to $a_j$):
        \begin{itemize}
            \item Load time: $s_j = a_j$.
            \item Compute time using $G^2$: $d_j = \varepsilon$, where $\varepsilon>0$ is a very small constant.
            \item Compute time using $G^1$: $p_j < d_j$.
        \end{itemize}
    \item For the additional block $0$:
        \begin{itemize}
            \item Load time $s_0 = 0$.
            \item Compute time using $G^1$: $p_0 = B$.
            \item Compute time using $G^2$: $d_0 > p_0$.
        \end{itemize}
    \item We ask whether there exists a schedule such that the overall latency $T$ is not larger than the prefill completion time.
\end{itemize}


\noindent \textbf{Key properties}.
This instance thus satisfy:
\begin{itemize}[leftmargin=15pt]
  \item For each chunk, block $0$ follows blocks $1, \dots, m$ and occupies $p_0 = B$ units of prefill time. Thus, in chunk~1 and chunk~2 we obtain two disjoint windows of length $B$ each. Any switching placed inside one of these windows does not extend the prefill completion time. Conversely, placing a switch outside these windows delays completion beyond $T$.
  \item Since all switchings must fit within these two windows of total length $2B$, the only way to complete them before prefill ends is to divide them into two groups of total duration $B$ each. This is exactly the Partition problem.
\end{itemize}

\noindent\textbf{Correctness.}  
\begin{itemize}[leftmargin=15pt]
\item \textbf{($\Rightarrow$)}  
If the Partition instance is solvable, i.e., there exists a subset $I$ such that $\sum_{j \in I} a_j = B$, then in the scheduling instance we place all switchings for indices in $I$ inside the window of chunk~1 (block $0$), and place the rest in chunk~2's window. Each window contains exactly $B$ switching time, all of which finish before the prefill completion time. Hence, this scheduling satisfies that the overall latency is not larger than the prefill completion time.

\item \textbf{($\Leftarrow$)}  
If the scheduling instance admits a feasible schedule with overall latency $\le T$, then all switchings must be finished before the prefill total time $T$. The only available periods where switchings can be hidden without delaying prefill are the two $B$-length windows (one in each chunk at block $0$). Since switchings are serialized, their total time $2B$ must be exactly distributed across the two windows. This requires splitting the switchings into two groups summing to $B$ each, which directly provides a solution to the Partition instance.

Thus, we have equivalence:
\[
\text{Partition is solvable} \;\;\Longleftrightarrow\;\; \text{Scheduling instance is solvable}.
\]

\end{itemize}

\noindent\textbf{Conclusion.} The reduction can be performed in polynomial time. Since Partition is NP-complete, the prefill scheduling problem is NP-hard. 

\end{document}